\documentclass[letterpaper, 10 pt, conference]{ieeeconf}  %
\IEEEoverridecommandlockouts      

\usepackage{times}
\usepackage[pdftex]{graphicx}
\usepackage{graphics}
\usepackage[cmex10]{amsmath}
\usepackage{url}

\usepackage{array}
\usepackage{tabularx}
\usepackage{multicol}
\usepackage[pagebackref=true,breaklinks=true,letterpaper=true,colorlinks,bookmarks=false]{hyperref}
\usepackage{amsmath}
\usepackage{graphicx} 
\usepackage{times}
\usepackage{booktabs}
\usepackage{epstopdf}
\usepackage{rotating}
\usepackage{csquotes}

\usepackage{amssymb}
\usepackage{amsmath}
\usepackage[all]{xy}    
\usepackage{ifthen}
\usepackage[usenames,dvipsnames]{color}
\usepackage{enumerate}
\usepackage{bm}
\usepackage{siunitx}
\usepackage{xargs}                      

\usepackage{caption}
\usepackage{    subfigure}
\usepackage[font=footnotesize]{caption}
\usepackage{color}
\newcommand{\bma}[1]{\left[\begin{array}{#1}}
\newcommand{\ema}{\end{array}\right]}
\usepackage{graphicx,dblfloatfix}
\usepackage[english]{babel}
\usepackage{blindtext}

\usepackage[pdftex,dvipsnames]{xcolor}  
\usepackage[colorinlistoftodos,prependcaption,textsize=tiny]{todonotes}
\newcommandx{\change}[2][1=]{\todo[linecolor=blue,backgroundcolor=white!25,bordercolor=blue,fancyline,#1]{#2}}
\newcommandx{\unsure}[2][1=]{\todo[linecolor=red,backgroundcolor=red!25,bordercolor=red,#1]{#2}}
\newcommandx{\info}[2][1=]{\todo[linecolor=OliveGreen,backgroundcolor=OliveGreen!25,bordercolor=OliveGreen,#1]{#2}}
\newcommandx{\improvement}[2][1=]{\todo[linecolor=Plum,backgroundcolor=Plum!25,bordercolor=Plum,#1]{#2}}
\newcommandx{\thiswillnotshow}[2][1=]{\todo[disable,#1]{#2}}

\makeatletter
\let\NAT@parse\undefined
\makeatother
\usepackage[numbers]{natbib}
\newcommand{\myparagraph}[1]{\vspace{0.07in}\noindent\textbf{#1}}

\title{\LARGE \bf
Tactile Regrasp: Grasp Adjustments via Simulated \\ Tactile Transformations
}

\author{Francois R. Hogan$^{*}$, Maria Bauza$^{*}$, Oleguer Canal, Elliott Donlon, and Alberto Rodriguez
\\ Department of Mechanical Engineering ---
    Massachusetts Institute of Technology \\ \tt\small $<$fhogan,bauza,oleguer,edonlon,albertor$>$@mit.edu  
    \thanks{This work was funded by the Amazon Research Awards and the NSF award [IIS-1637753] through the National Robotics Initiative.}
    \thanks{$^\star$ Authors with equal contribution.}
    }

\begin{document}

\maketitle
\thispagestyle{empty}
\pagestyle{empty}

\begin{abstract}

This paper presents a novel regrasp control policy that makes use of tactile sensing to plan local grasp adjustments. 
Our approach determines regrasp actions by virtually searching for local transformations of tactile measurements that improve the quality of the grasp.

First, we construct a tactile-based grasp quality metric using a deep convolutional neural network trained on over 2800 grasps. The quality of each grasp, a continuous value between 0 and 1, is determined experimentally by measuring its resistance to external perturbations.
Second, we simulate the tactile imprints associated with robot motions relative to the initial grasp by performing rigid-body transformations of the given tactile measurements. The newly generated tactile imprints are evaluated with the learned grasp quality network and the regrasp action is chosen to maximize the grasp quality.

Results show that the grasp quality network can  predict the outcome of grasps with an average accuracy of 85$\%$ on known objects and 75$\%$ on novel objects. The regrasp control policy improves the success rate of grasp actions by an average relative increase of 70$\%$ on a test set of 8 objects. We  provide a video summarizing our approach at \href{https://youtu.be/gjn7DmfpwDk}{https://youtu.be/gjn7DmfpwDk}.

\end{abstract}

\section{Introduction}

In this paper we demonstrate that a robot can use tactile sensing to improve the performance of a grasping system.
%
%
In particular we propose to use tactile information to assess in real-time the quality of a grasp, predict grasp failures, and simulate local tactile transformations to search for a grasp adjustment that improves its quality.

Our long term goal is to enable the use of tactile feedback in robotic manipulation and grasping. The lack of tactile reasoning is one of the main limitations of robotic grasping and a long-standing challenge in the robotics community. After decades of advances in sensing instrumentation and processing power, the basic question remains: \emph{How should robots make use of sensed contact information?}

%
For decades, grasping has been approached as a geometric problem, where object pose and shape are used to plan a grasp.
In that approach, the quality of a grasp is modeled directly as the geometric fit between gripper and object, exploiting computational approximations like point contacts or simple friction models~\cite{bicchi2000robotic}.

\begin{figure}[t]
\centering
	\includegraphics[width=7cm]{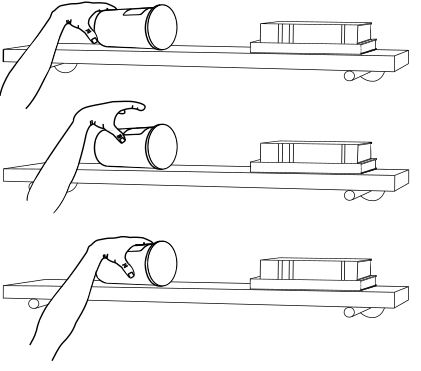}
\centering
\caption{Illustration of a regrasp action using tactile feedback. In the absence of  visual information, humans are  capable of performing dexterous grasping actions. The animation depicts a human trying to grasp an object that is outside its range of vision. Using only tactile information, the hand is moved to secure a better grasp on the object. Our approach is motivated by this example and explores how grasping can benefit from making local grasp adjustments based on tactile  information. } \label{fig:tactile_reflex}
\end{figure}

Current research efforts, boosted by advances in learning, and competitions like the Amazon Robotics Challenge~\cite{Correll2016}, have converged to a more integrated approach that plans grasps by finding affordances directly in an RGBD image.
This approach has increased the robustness of grasping systems, but is limited in two ways: 1) by the realism and generalization of the affordance models, either learned from experience~\cite{pinto2016, zeng_2017} or from simulation~\cite{mahler2017dex}; and 2) by the open-loop/eyes-closed execution of the grasping behavior. This is specially limiting in scenarios with occlusions. Figure~\ref{fig:tactile_reflex} depicts an example where a human attempts to grasp a jar from an overhead shelf. In the absence of visual information, humans rely on tactile information to adjust their hand position and secure a grasp.

    

%

We propose an approach to tactile regrasp based on two main contributions:

\begin{enumerate}
    \item \textbf{Grasp Quality Metric}. Self-Supervised approach to learn a tactile-only metric of grasp quality. This metric, a continuous value between 0 and 1, evaluates how likely a given grasp is to resist external forces emulated by shaking the robot gripper.
    \item \textbf{Regrasp Policy}. Planning grasp adjustments via simulating local tactile transformations. We use the tactile-based grasp quality metric to search for grasp adjustments that better secure the object grasp.
\end{enumerate}
%
%
\begin{figure}[t]
\centering
{
		\includegraphics[width=8.cm]{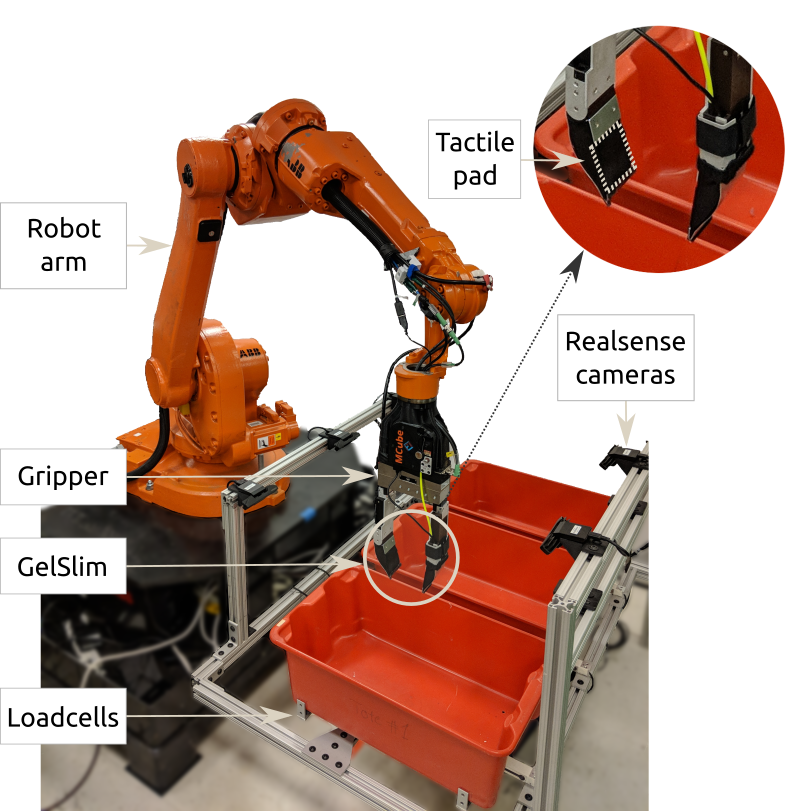} %
}
\centering
\caption{Setup used for autonomous grasping. The system is equipped with a variety of sensors such as load cells, RGBD cameras, and tactile sensors. The design of the system builds on team MIT-Princeton's robotic solution for the Amazon Robotics Challenge 2017 \cite{zeng_2017} and is adapted to perform autonomous and reliable grasps over long periods of time  without any human supervision.} \label{fig:system_setup}
\end{figure}

\begin{figure*}[t]
\centering
{
		\includegraphics[width=\linewidth]{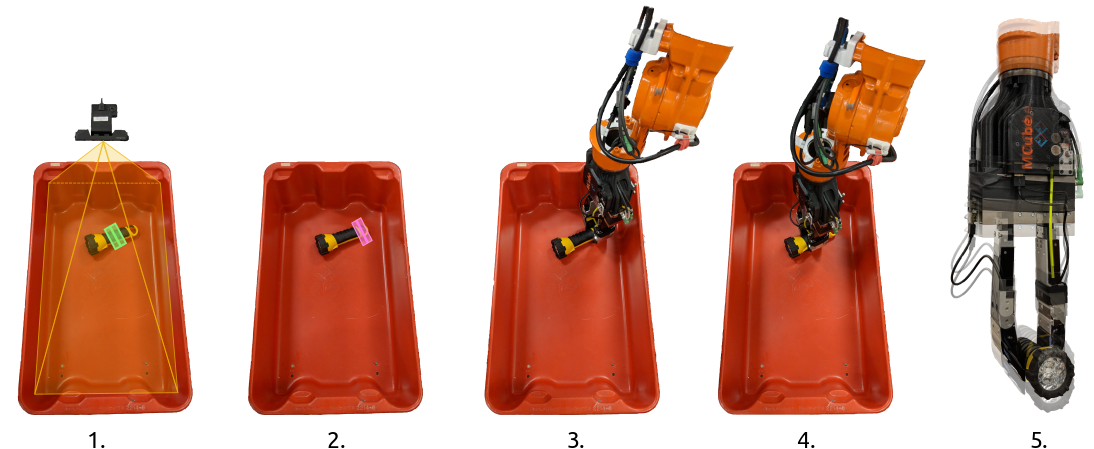} %
}
\centering
\caption{Grasping approach overview. 1) We perceive the 3D pointcloud of the scene and compute an antipodal grasp affordance. 2) We modify the initial proposal to mimic a more realistic grasping system where errors and perturbations can occur. 3) The robot descends to the planned grasp, closes its fingers and records the tactile imprints. 4) Based on the tactile imprints, a local adjustment is planned and the robot performs the associated regrasp action. 5) The robot shakes the gripper to evaluate the quality of the regrasp action. } \label{fig:approach}
\end{figure*}
One can think of this approach as a hybrid between learning and modeling.
Our approach is learning-based, because we build and validate the grasp metric directly from data, for which we collect more than 2800 grasps with the robotic system shown in Fig.~\ref{fig:system_setup}. This metric is specially apt at exploiting the geometry, compliance, friction, and sensing capabilities of the gripper we use.
Our method is also model-based because rather than learning directly a regrasp policy tailored to the specifics of the objects and scenario where the data was collected, we plan regrasps using the learned grasp quality model. 
We believe this distinction is important for the generalization power of the algorithm.



Results show that the model-based regrasp policy, guided by the data-driven grasp quality metric, in average increases the relative accuracy  of  grasping by $70 \%$ on a test set of $8$ objects.


\section{Related Work}

Grasp planning has a long and rich history in robotic manipulation \cite{bicchi2000robotic, borst_2004, berenson_2007, ciocarlie_2009, sahbani_2012}. Conventional approaches focus on determining grasp  configurations that ensure  there is some form of geometric closure on the object. The success of such grasp planning strategies relies on accurate state estimation of the object pose. Moreover, it has been shown that classical grasp metrics are weak predictors of grasp quality when implemented on a
physical robotic platform \cite{goins_2014}. 

Recently, a large portion of the robotic manipulation community has converged to grasp planning methods that are agnostic to the identity and state of the object. Some model-based approaches rank grasp points according to a grasp quality metric that is based on local properties of the camera point cloud~\cite{ten2015using, gualtieri2017}. Similarly, leveraging recent advances in computer vision and deep learning, many researchers have turned to data-driven methods that localize grasp points directly from the RGBD image \cite{redmon2015,  pinto2016, pinto2017, mahler2017dex, levine2016}. Both of these object agnostic approaches have led to effective grasp planning algorithms that deal with a large variety of object types and dense clutter. However, one limitation of such approaches is that they are most often implemented in an open-loop fashion, where the robot motion remains unchanged after the initial grasp location is determined. Due to inaccuracies in perception, dense clutter, and unaccounted object motions, the robot often encounters unanticipated events, such as premature contact, collisions with other objects, or imperfect grasp point locations. 

To address these issues, there have been recent attempts to develop closed-loop approaches to grasping. In \cite{levine2016}, a deep reinforcement learning approach is used to learn a closed-loop control policy from RGB video feed. In \cite{luberto_2017}, RGBD cameras are combined with infrared sensors to design a reactive algorithm that improves the robustness and adaptability of grasps of unknown objects with uncertain position. In \cite{viereck_2017}, sensor data is generated in simulation to build a reactive grasping policy that computes grasp affordances based on depth images.

In an effort to enable more effective closed-loop approaches to robotic grasping,  researchers have turned to tactile sensing to enable reactive behavior based on what the robot feels rather than what it sees. As reviewed in \cite{youssef_2011}, there is a wealth of literature concerning the use of tactile sensors in robotic manipulation. Tactile sensors have already proved effective at detecting contact slip between the gripper and grasped objects \cite{stachowsky_2016, ajoudani_2016, dong_2017}, estimating contact forces~\cite{maria_2012}, and localizing objects~\cite{klingensmith_2016, izatt_2017}. In \cite{bekiroglu_2013}, a grasp quality predictor is constructed using self-supervised learning to predict the probability of success given tactile information. In \cite{chebotar_2016}, a reinforcement learning approach  uses a grasp quality predictor to learn grasp adjustments for a cylindrical object based on tactile feedback. In \cite{pastor2012towards}, tactile sensors are integrated into the Dynamic Motion Primitives (DMP) framework to enable Associative Skill Memories (ASM). This imitation learning technique allows a robotic manipulator to replicate both the kinematics of the robot along with the sensorimotor measurements it encounters during expert demonstrations. In \cite{hsiao_2010}, pressure sensors located in the robot fingers are used to adjust the planned trajectory of a robot in real time to improve the robustness of horizontal grasps.

In this research, we make use of GelSlim \cite{donlon2018gelslim}, a tactile sensor based on Gelsight \cite{yuan_2017}. The Gelsight sensor has proved useful to identify object properties \cite{li_2013, yuan_2017_v2}, slippage detection \cite{dong_2017}, object localization \cite{li_2014, izatt_2017}, and grasp stability evaluation. We are particularly inspired by \cite{calandra_2017}, who showed that a combination of tactile sensing with visual information can reliably determine whether a given grasp will lead to a successful execution. In this research, we focus on the case of pure tactile feedback, where only local contact information is used for controller design.

\section{Approach}
\label{sec:approach}

This section summarizes our approach to design a tactile-based policy for grasp adjustment. First, we learn a tactile-based model of the quality of a grasp. Rather than relying on classical grasp metrics, we make use of a data-driven approach that better captures the physical properties of our particular gripper and tactile sensor. Next, we exploit this metric of grasp quality to plan regrasp actions.

To achieve this goal, we design a policy that makes local grasp adjustments after having made initial contact with the object. Figure~\ref{fig:approach} illustrates the approach used to implement and evaluate the policy. First, an antipodal grasp affordance, that can come from any grasp planner, is computed from the perceived 3D pointcloud. In this work we consider the grasp baseline algorithm used by the MIT-Princeton team at the ARC 2017 \cite{zeng_2017}. Second, we introduce noise to the grasp affordance to simulate disturbances that can worsen the grasp proposals. Third, the robot descends to the modified grasp location, closes its fingers and records the tactile imprints.  Fourth, based on the tactile imprints, a local adjustment is planned and the robot performs the associated regrasp action. Finally, the robot shakes the gripper to evaluate the quality of the regrasp adjustment.

We break down this problem and organize the paper as below:
\begin{itemize}
    \item \textbf{Robot System}. Describes the robot platform used for data collection and policy evaluation (Section~\ref{sec:robotic_system}).
    \item \textbf{Data Collection}. Summarizes the experimental procedure used to collect ground truth labels of grasp quality from tactile imprints (Section~\ref{sec:data_collection}).  
    \item \textbf{Tactile Metric on Grasp Quality}. Details the neural network architecture used to infer a grasp quality metric from collected data (Section~\ref{sec:grasp_stability}).
    \item \textbf{Grasp Adjustment}. Introduces a tactile-based reactive regrasp policy by planning grasp adjustments (Section~\ref{sec:regrasp_strategy}). 
    \item \textbf{Experimental Results}. Assessment of the quality of the learned metric and the regrasp policy on a set of experiments conducted on known and unknown objects (Section~\ref{sec:experimental_results}).
\end{itemize}

\section{Robotic System}
\label{sec:robotic_system}

This section presents the main components of our autonomous system for robotic grasping. We first describe our robotic platform and then briefly review the tactile sensor used to detect contact interactions during grasping.

\myparagraph{System setup.} Our system setup shown in Fig.~\ref{fig:system_setup} features a 6DOF industrial robot arm (ABB IRB 1600id), equipped with three grasping stations.  The robot's end-effector comprises of a  parallel jaw gripper (WSG-50 Weiss) on which two custom built fingers are mounted. Each finger integrates a high resolution image-based tactile sensor. Each grasping station includes a storage bin and two statically-mounted RGBD cameras (RealSense SR300). 
Underneath each bin are installed two load cells (Loadstar RSP1) used to detect when contact is made between the robot end-effector and the bin. This has proved crucial to ensure safe executions while autonomously collecting data with a position controlled industrial arm.  

\begin{figure}[h]
\centering
{
		\includegraphics[width=\linewidth]{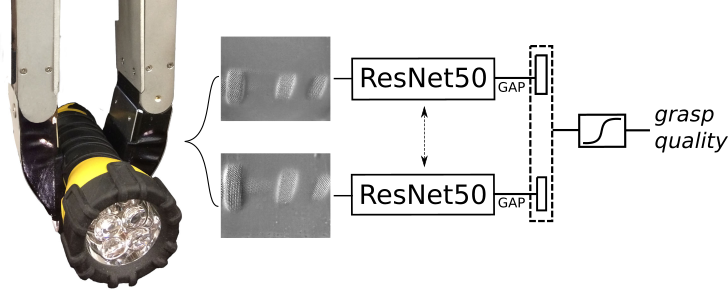} %
}
\centering
\caption{Tactile sensor and network architecture used to predict grasp quality. The learner takes as input the two RGB tactile images obtained right before lifting the object. Both are processed using a shared ResNet50 architecture pre-trained on the ImageNet dataset. We crop the ResNet50 architecture at the GAP layer and concatenate the results from both images into a single layer. Finally we add a dense layer with a sigmoid activation function to output the quality of the grasp.} \label{fig:network_architecture}
\end{figure}

\myparagraph{GelSlim Sensor.} The GelSlim sensor \cite{donlon2018gelslim}  in Fig.~\ref{fig:network_architecture} is an optical-based tactile sensor inspired from Gelsight \cite{GelSight_review} that renders high resolution images of the contact surface geometry. When an object is pressed against the GelSlim, a camera located inside the finger of the gripper captures the deformation of the sensor's membrane. Figure~\ref{fig:network_architecture} shows an example of the image-based output of the sensor when grasping an object with a ridged surface (\textit{flashlight}).

\section{Data Collection}
\label{sec:data_collection}
We use a self-supervised learning approach to collect a dataset of tactile signatures labelled with their grasp quality. We self-supervise the system by:

\begin{enumerate}
    \item \textbf{Predicting grasp affordances}. To obtain good initial grasps affordances, our system relies on the vision-based solution proposed in \cite{zeng_2017}, which proved successful at grasping a large range of objects in clutter. 
    Grasp affordances are selected by identifying antipodal grasps from the point cloud, i.e., regions where the gripper fits and is likely to have something between its fingers. 

    \item \textbf{Noise infusion}. To  build a reliable metric for grasp quality, we aim to collect data that resembles realistic scenarios where incomplete visual information, noise in the robot dynamics or unaccounted object motions degrade the grasping performance. To emulate this, we add noise to the grasp proposal provided by the grasp planner in the lateral direction to the gripper (normal direction of the plane formed by the two fingers). The magnitude of the noise is tailored to the object dimensions to encourage grasps at the borders while maintaining a similar ratio between successes and failures. These are grasps for which the outcome is more challenging to predict leading to a more even distribution between successful and failed grasps.
    
    \item \textbf{Grasp evaluation}.
%
%
%
%
%
%

The robot assesses the quality of a grasp by shaking its end-effector after grasping an object with a constant force of 30N. 
The grasp quality is proportional to the time the grasp resists the shake:
\begin{equation}\label{eq:score}
\begin{cases} 
      0 & \text{if \text{ failure occurs before $t_{0}$}}, \\
      0.5 \cdot \frac{ t_{i}-t_{0}}{T_{s}} & \text{if \text{ failure occurs at $t_{i}$}}, \\
      1 & \text{if \text{ no failure is detected}},
   \end{cases}
\end{equation}
where $t_{0}$, and $t_{i}$ denote the shaking starting time and the detected impact time after the object falls. The term $T_{s}$ represents the total robot shaking time and has been set experimentally to 4 seconds. Note that the score for failed grasps is always lower or equal to 0.5 emphasizing the difference between successful and failed grasps.

\end{enumerate}

To evaluate our method, we collect grasps from $20$ objects with different shapes and textures, as reported in Tables~\ref{table:cross-validation} and~\ref{table:regrasp_accuracy}. Our dataset includes diversity in object shape (cylindrical, spherical, square, etc.) as well as material texture and hardness, which produces a varied set of imprints on the tactile sensor.

\section{Tactile-Based Grasp Quality}
\label{sec:grasp_stability}
In this section we describe the construction of a grasp quality metric--how likely a grasp is to succeed--given only tactile information. 
By making use of the grasp quality metric, a robot picking system can reason about the state of the grasp in real time, detect and prevent failures, and plan regrasp actions as proposed in Section \ref{sec:regrasp_strategy}. 

Given a set of grasps labelled as in Eq. \eqref{eq:score} and their respective tactile measurements, we learn a model that predicts how likely an object is to resist external perturbations. We formulate this problem as a self-supervised task and directly learn a regressor that outputs the quality of a grasp. Exploiting the image-based nature of the tactile sensor, we make use of deep convolutional neural networks to learn the grasp quality metric.  

\myparagraph{Network architecture.} We select the network's input to be the two tactile images captured prior to lifting the object. 
The images are processed using a ResNet50 architecture \cite{he2016} with pretrained weights from ImageNet as shown in Fig.~\ref{fig:network_architecture}. The outputs from the ResNet  are concatenated together and passed through a sigmoid output layer to assess the grasp quality. The resulting neural network is trained using the Adam optimizer and the cross-entropy loss function. The experimental evaluation of the metric is detailed in Section~\ref{subsec:cross_validation}.




\begin{figure*}[t]
\centering
{
	 \vspace{5mm}	\includegraphics[width=\linewidth]{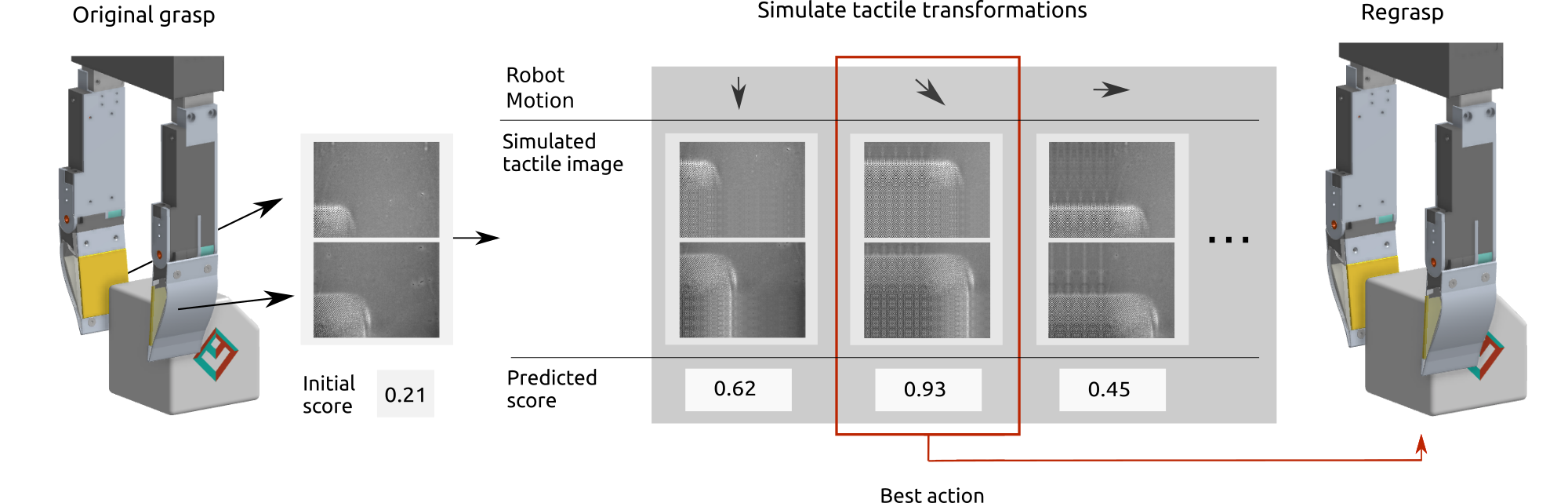} %
}
\centering
\caption{\textbf{Regrasp control policy.} First, the robot closes its fingers and the resulting tactile imprints are processed to determine the grasp quality. Second, we generate possible motions of the robot gripper by simulating the tactile images that would be obtained after each regrasp. Third, we label each simulated tactile image using our trained grasp-stability metric and select the ones that achieve the highest score. Finally, we perform the regrasp action associated with the best ranked tactile images.} \label{fig:control_policy}
\end{figure*}

\section{Grasp Adjustment Policy}\label{sec:regrasp_strategy}

Given a tactile image, how can we determine local gripper adjustments that will improve the quality of a grasp? This section introduces a tactile-only regrasp policy that exploits the grasp quality metric described in Sec.~\ref{sec:grasp_stability}.




The proposed approach, illustrated in Fig.~\ref{fig:control_policy}, follows these steps: 
\begin{enumerate}
    \item Capture the tactile image from the initial grasp.
    \item Explore candidate regrasps along with their simulated tactile images.
    \item Assign grasp quality scores to the simulated tactile images.
    \item Perform the regrasp action associated with the highest ranked tactile transformation.
\end{enumerate}
%
%
The example robot motions shown in Fig.~\ref{fig:control_policy} correspond to the robot moving down, right-down, and right. The gripper actions are taken in the plane defined by the parallel jaw gripper. 
Given a particular gripper motion, we compute its associated tactile imprint by using rigid-body transformations of the initial tactile images. In this work we only consider translations of the tactile images although more complex transformations like rotations or warping could also be considered. 
Motions between the image frame and the world frame are related linearly as:
\begin{equation}
\bma{cc}
     x_{pixel}\\y_{pixel}
\ema
=
\bma{cc}
     \frac{r_x}{w_{g,x}}x_{hand}\\  \frac{r_y}{w_{g,y}}y_{hand}
\ema,
\end{equation}
where $r$ and $w_{g}$ denote the resolution and width of the tactile sensor, respectively. The subscripts \textit{pixel} and \textit{hand} represent the coordinate reference frames of the image and the robot hand. 
After translating an image in pixel space, we perform a mirror process to populate the newly introduced pixels. The mirror considered has a width of 15 pixels and is continuously applied until the entire image is covered.  This technique is best suited for objects that are wider than the width of the tactile sensor. 

A feature of this method is that the policy is agnostic to the object's identity and global geometry, and provides feedback uniquely based on the local geometry of the contact patch recorded during the initial contact. The proposed policy is based on two approximations:
\begin{enumerate}
    \item The pose of the object remains unchanged between two grasps.
    \item The tactile imprint recorded after the regrasp action is a rigid body transformation of the original tactile image.
\end{enumerate}

Due to the nature of the parallel jaw gripper, the first approximation is most often satisfied as long as the object is resting in a stable configuration. The second approximation is more likely to be violated, as the regrasp action will introduce an unknown contact area to the tactile sensor. In practice, we observe that the policy performs well even when these assumptions are violated.

\begin{table*}[!t]
  \centering
  \setlength{\tabcolsep}{4 pt}
  \vspace{5mm}
  \caption{Grasp quality accuracy   on unknown objects}
  \begin{tabular}{c|c|c|c|c|c|c|c|c |c|c|c}
         air & blue & conair & dawn & dog & elmer's &glucose & dorcy & extension  &  staples & speed & tennis \\
          freshener & tube & brush & soap & bone & glue & bottle & flashlight & cord & box & stick & container \\
          ($215$ g) & ($56$ g)  & ($58$ g)  &  ($113$ g)  & ($172$ g)  & ($146$ g)  & ($274$ g) & ($293$ g) & ($137$ g)  & ($158$ g) &  ($273$ g)  & ($223$ g)    \\
          \hline
         \hspace{-3mm}
    \begin{minipage}{0.08\textwidth}
    \centering
    \vspace{2mm}
        \includegraphics[width=\linewidth, height=17mm, keepaspectratio]{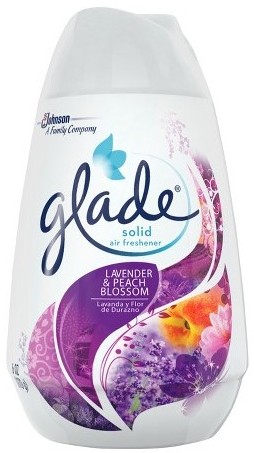}
        \vspace{2mm}
    \vspace{2mm} \end{minipage}
    \hspace{-3mm}
    &
    \hspace{-3mm}
    \begin{minipage}{0.08\textwidth}
    \centering
    \vspace{2mm}
        \includegraphics[width=\linewidth, height=14mm, keepaspectratio]{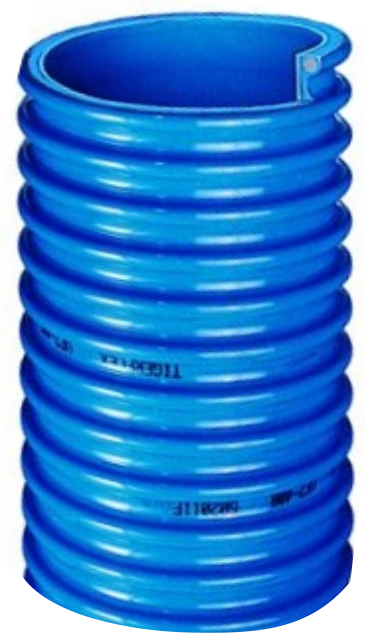}
    \vspace{2mm} \end{minipage}
    \hspace{-3mm}
    &
    \hspace{-3mm}
    \begin{minipage}{0.08\textwidth}
    \centering
    \vspace{2mm}
      \includegraphics[width=\linewidth, height=17mm, keepaspectratio]{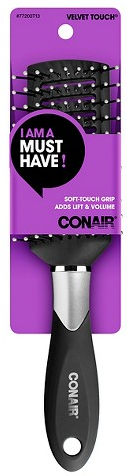}
    \vspace{2mm} \end{minipage}
    \hspace{-3mm}
    &
    \hspace{-3mm}
    \begin{minipage}{0.08\textwidth}
    \centering
    \vspace{2mm}
        \includegraphics[width=\linewidth, height=14mm, keepaspectratio]{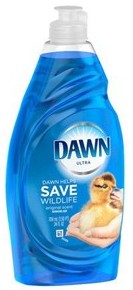}
    \vspace{2mm} \end{minipage}
    \hspace{-3mm}
    &
    \hspace{-3mm}
    \begin{minipage}{0.08\textwidth}
    \centering
    \vspace{2mm}
        \includegraphics[width=\linewidth, height=17mm, keepaspectratio]{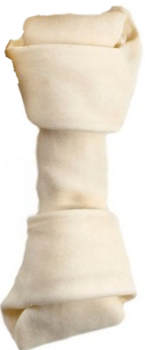}
    \vspace{2mm} \end{minipage}
    \hspace{-3mm}
    &
    \hspace{-3mm}
    \begin{minipage}{0.08\textwidth}
    \centering
    \vspace{2mm}
        \includegraphics[width=\linewidth, height=17mm, keepaspectratio]{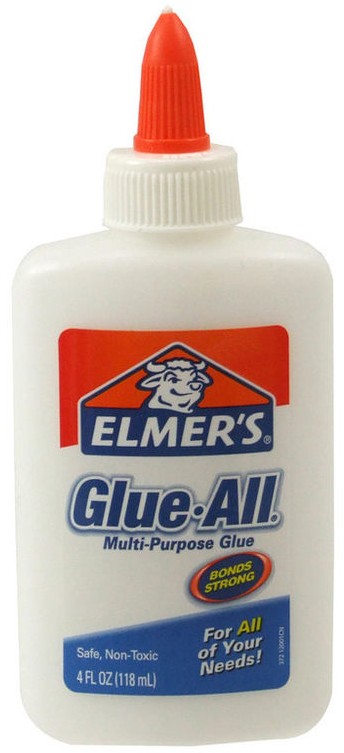}
    \vspace{2mm} \end{minipage}
    \hspace{-3mm}
    &
    \hspace{-3mm}
    \begin{minipage}{0.08\textwidth}
    \centering
    \vspace{2mm}
        \includegraphics[width=\linewidth, height=17mm, keepaspectratio]{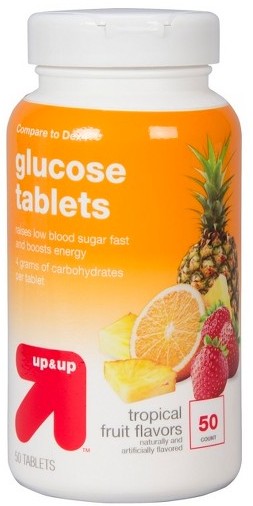}
    \vspace{2mm} \end{minipage}
    \hspace{-3mm}
    &
    \hspace{-3mm}
    \begin{minipage}{0.08\textwidth}
    \centering
    \vspace{2mm}
        \includegraphics[width=\linewidth, height=17mm, keepaspectratio]{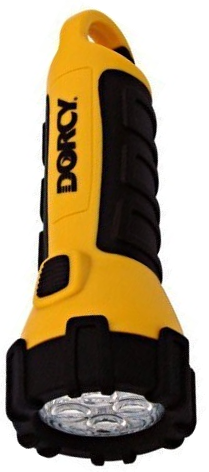}
    \vspace{2mm} \end{minipage}
    \hspace{-3mm}
    &
    \hspace{-3mm}
    \begin{minipage}{0.08\textwidth}
    \centering
    \vspace{2mm}
        \includegraphics[width=\linewidth, height=16mm, keepaspectratio]{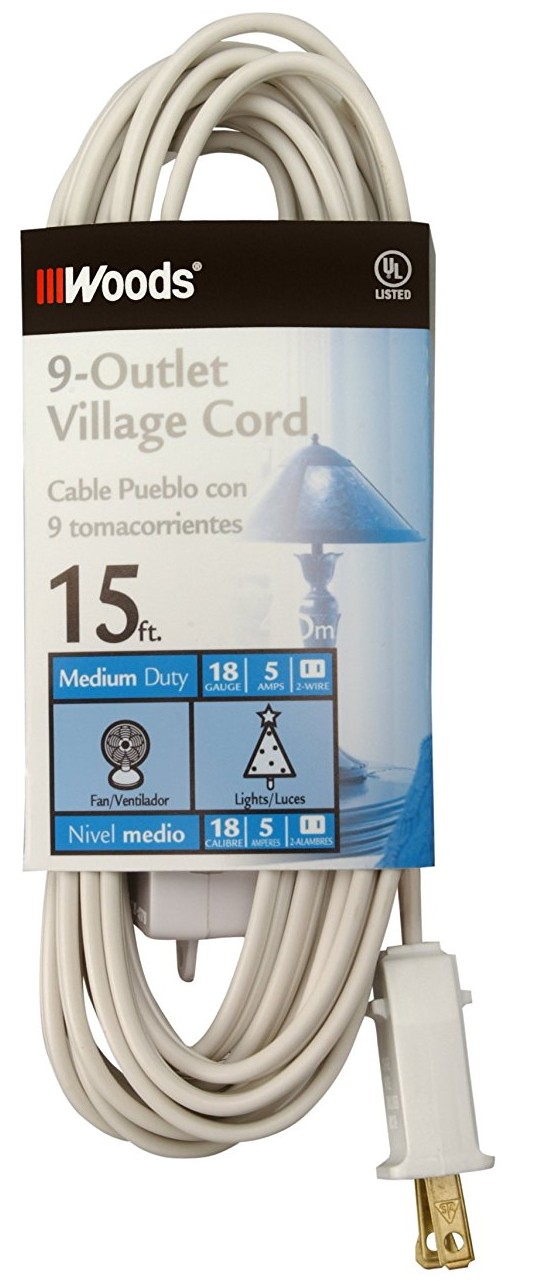}
    \vspace{2mm} \end{minipage}
    \hspace{-3mm}
    &
    \hspace{-3mm}
    \begin{minipage}{0.08\textwidth}
    \centering
    \vspace{2mm}
        \includegraphics[width=\linewidth, height=17mm, keepaspectratio]{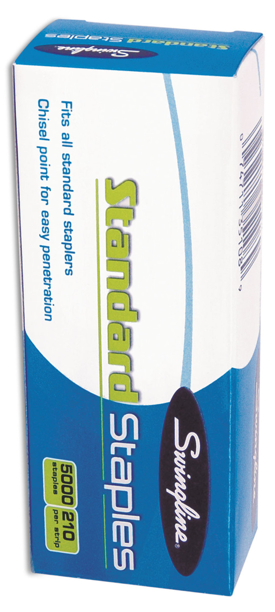}
    \vspace{2mm} \end{minipage}
    \hspace{-3mm}
    &
    \hspace{-3mm}
    \begin{minipage}{0.08\textwidth}
    \centering
    \vspace{2mm}
        \includegraphics[width=\linewidth, height=17mm, keepaspectratio]{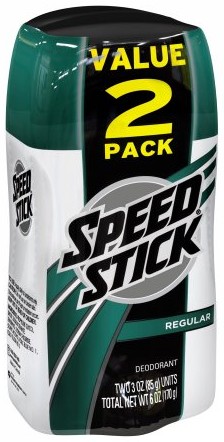}
    \vspace{2mm} \end{minipage}
    \hspace{-3mm}
    &
    \hspace{-3mm}
    \begin{minipage}{0.08\textwidth}
    \centering
    \vspace{2mm}
        \includegraphics[width=\linewidth, height=17mm, keepaspectratio]{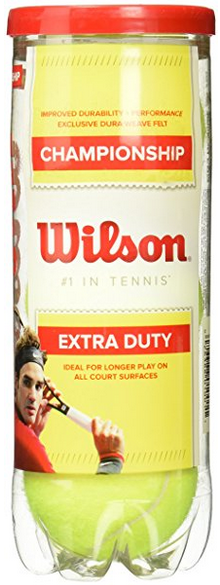}
    \vspace{2mm} \end{minipage}
    \\ \hline \rule[-1ex]{0pt}{3.3ex}
    77 $\%$ & 96 $\%$ & 49 $\%$& 63 $\%$& 66 $\%$& 84 $\%$& 61 $\%$  & 82 $\%$& 79 $\%$& 91 $\%$ & 73 $\%$& 79 $\%$ \\\hline 
  \end{tabular}
  \label{table:cross-validation}
  \vspace{-3mm}
\end{table*}

\section{Experimental Results}\label{sec:experimental_results}

In this section, we  evaluate the grasp quality metric learned in Sec.\ref{sec:grasp_stability} by performing cross validation on a set of 12 objects, and analyze the performance of the regrasp policy proposed in Sec.\ref{sec:regrasp_strategy} on a set of 8 new objects.

\subsection{Grasp quality metric}
\label{subsec:cross_validation}

We assess the performance of the grasp quality predictions in two different cases: known and unknown objects. 

\myparagraph{Performance on known objects.} We collect over 300 grasps with their individual quality scores for each of the 12 objects in Table~\ref{table:cross-validation}\footnote{With the exception of the item \textit{air freshener} that broke prematurely}, and split them into two sets: training (~2800 grasps) and testing (~200 grasps). Setting the decision boundary at 0.5 for successful grasps, we obtain a training accuracy of 90$\%$ and a testing accuracy of 85$\%$. These results show that given a new grasp from a known object we can reliably infer its grasp quality.
 

\begin{figure*}[!b]
\centering
{
		\includegraphics[width=16.cm]{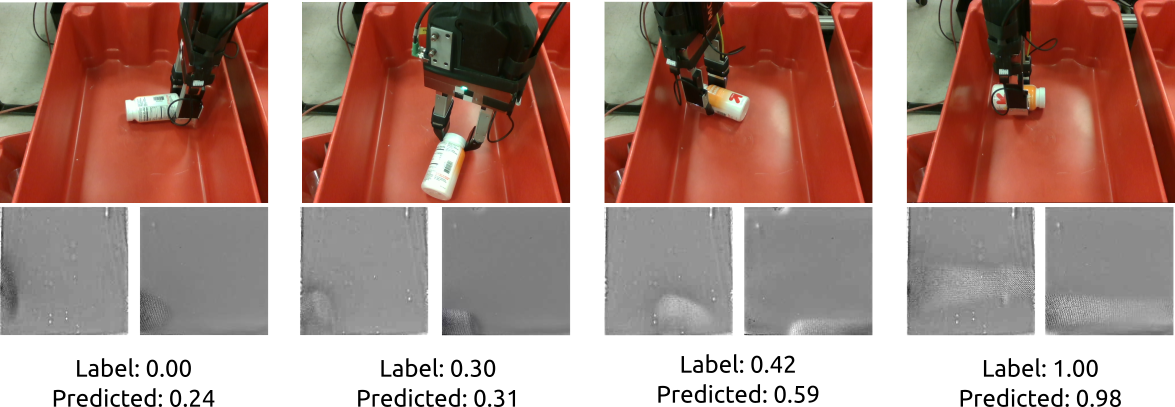} %
}
\centering
\caption{\textbf{Qualitative analysis of the learned grasp metric.} Each image shows a different grasp from the test set with its associated quality (label) and its predicted quality given the tactile imprints obtained during grasping. We observe that our grasp quality model can recover the underlying distribution of labels and discriminate better the quality scores than a simple 0-1 classification task.} \label{fig:grasp_scores}
\end{figure*}

\myparagraph{Performance on unknown objects.} To evaluate the performance of the grasp quality metric on unknown objects, we use a cross-validation strategy using the grasps from the objects in Table~\ref{table:cross-validation}. We train 12 different object specific grasp quality metrics by withholding the grasps from one object for validation and training each model using the data from the remaining 11 objects\footnote{For fair comparison, we report the validation accuracy using a fixed number of 100 epochs to train each model.}.
%

%
The accuracy obtained for each object is shown in Table~\ref{table:cross-validation}. We obtain an average accuracy on unknown objects of 75$\%$, suggesting that our learned grasp quality metric has the ability to generalize to unknown objects. The disparity in accuracy between the different objects in Table~\ref{table:cross-validation} shows that the oddly-shaped and heavy objects are the most challenging to predict. For complex shapes, it is intuitive that the performance decreases as their tactile imprints differ from the training distribution. Such is the case of \textit{conair brush} and \textit{dog bone} that have very particular geometries.  
For heavy objects, we notice that similar imprints are likely to be labelled with different grasp qualities, as small differences in the grasp position might have a significant impact on the quality of the grasp.  

Figure~\ref{fig:grasp_scores} shows examples of grasps with high and low quality together with the quality score predicted by our model. From these instances, we observe that tactile information carries significant information relative to the nature of the grasp and is often sufficient to infer its quality.

\subsection{Grasp adjustment based on tactile information}
\label{subsec:regrasp_policy}

In situations of visual occlusion or large clutter, it is not possible to rely only on visual information to plan a successful grasp. In those cases, regrasp actions that consider tactile information can help boost the performance of a grasping system. In this work we explore the use of high resolution tactile information to improve the robustness of grasping systems.

We evaluate the regrasp policy in Sec.~\ref{sec:regrasp_strategy} with 8 new objects in Table~\ref{table:regrasp_accuracy}. These objects are selected to be challenging for regrasp actions due to their heavy weight or complex geometries. The regrasp strategy considers gripper motions that range from 0 to 3cm  in steps of 0.6cm in the normal direction of the gripper.  

In Table~\ref{table:regrasp_accuracy},  we show the accuracy averaged over 100 grasps both with and without grasp adjustments. In general the use of a regrasp action increases the overall accuracy by at least  14$\%$ and an average of approximately 24$\%$. If we compute the relative improvement, the average increase is 70$\%$. This high improvement is due to the ability of the tactile-based policy to deal with the diverse set of grasp proposals by correcting bad grasps and maintaining the good ones. 

\begin{figure*}[!b]
\centering
{
		\includegraphics[width=16.cm]{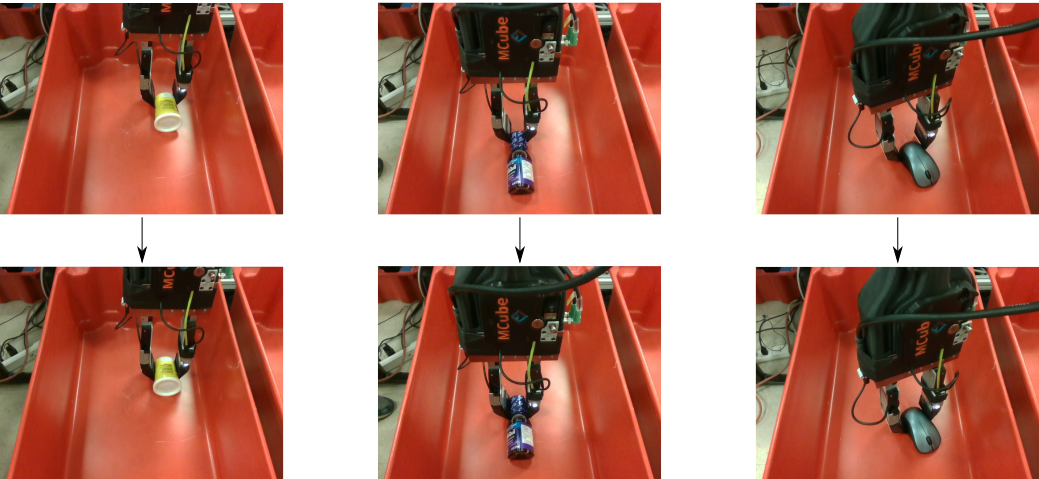} %
}
\centering
\caption{\textbf{Examples of regrasps using the tactile-based policy.} After making an initial grasp on the object, the gripper adjusts its position to improve the grasp quality of the tactile imprint. This re-positioning allows the robot to secure a better grasp on the object and improves the average  relative accuracy of the 8 new objects by 70$\%$.} \label{fig:grasp_policy_qualitative}
\end{figure*}

Figure~\ref{fig:grasp_policy_qualitative} shows different scenarios where the tactile-based policy is capable of improving the initial grasp affordances. Results show that robotic picking systems can benefit from using tactile information. In particular, tactile sensing can help during a grasp action to decide whether the grasp quality is good enough, or a regrasp action would be preferable.



\begin{table*}[!t]
  \centering
  \setlength{\tabcolsep}{4 pt}
  \vspace{5mm}
  \caption{Regrasp strategy improves grasping accuracy}
  \begin{tabular}{c|c|c|c|c|c |c|c|c}
      Objects  & black  & heavy  & logitech & metal & satin   &  soy & tomato  & ZzzQuil   \\
        &  pepper &   play-doh & mouse & bar &  care &  milk  & can & bottle  \\
         &  ($233$ g) &  ($332$ g) & ($123$ g) & ($255$ g) &  ($276$ g) & ($261$ g)  & ($367$ g) & ($237$ g) \\\hline
        
Images & 
 \hspace{-3mm}
    \begin{minipage}{0.1\textwidth}
    \centering \vspace{2mm}
        \includegraphics[width=\linewidth, height=14mm, keepaspectratio]{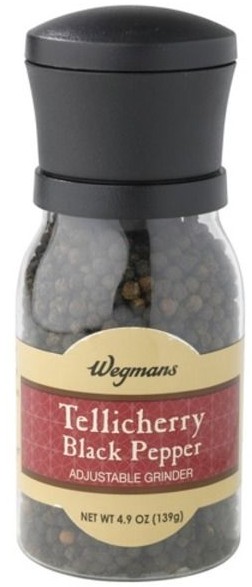}
    \vspace{2mm} \end{minipage}
    \hspace{-3mm}
    &
    \hspace{-3mm}
    \begin{minipage}{0.1\textwidth}
    \centering \vspace{2mm}
        \includegraphics[width=\linewidth, height=12mm, keepaspectratio]{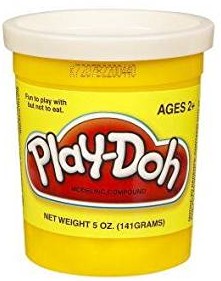}
    \vspace{2mm} \end{minipage}
        \hspace{-3mm}
    &
    \hspace{-3mm}
    \begin{minipage}{0.1\textwidth}
    \centering \vspace{2mm}
      \includegraphics[width=\linewidth, height=9mm, keepaspectratio]{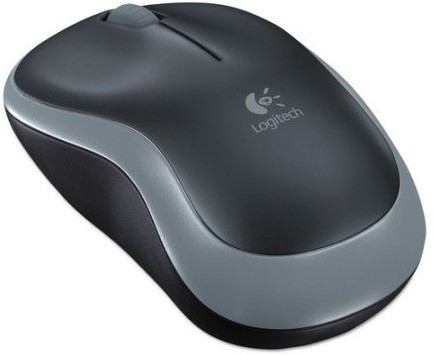}
    \vspace{2mm} \end{minipage}
        \hspace{-3mm}
    &
    \hspace{-3mm}
    \begin{minipage}{0.1\textwidth}
    \centering \vspace{2mm}
        \includegraphics[width=\linewidth, height=9mm, keepaspectratio]{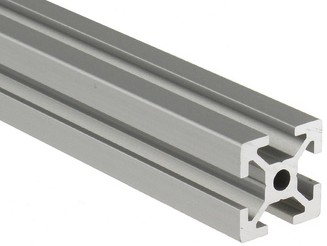}
    \vspace{2mm} \end{minipage}
        \hspace{-3mm}
    &
    \hspace{-3mm}
    \begin{minipage}{0.1\textwidth}
    \centering \vspace{2mm}
        \includegraphics[width=\linewidth, height=14mm, keepaspectratio]{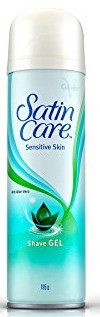}
    \vspace{2mm} \end{minipage}
        \hspace{-3mm}
    &
    \hspace{-3mm}
    \begin{minipage}{0.1\textwidth}
    \centering \vspace{2mm}
        \includegraphics[width=\linewidth, height=14mm, keepaspectratio]{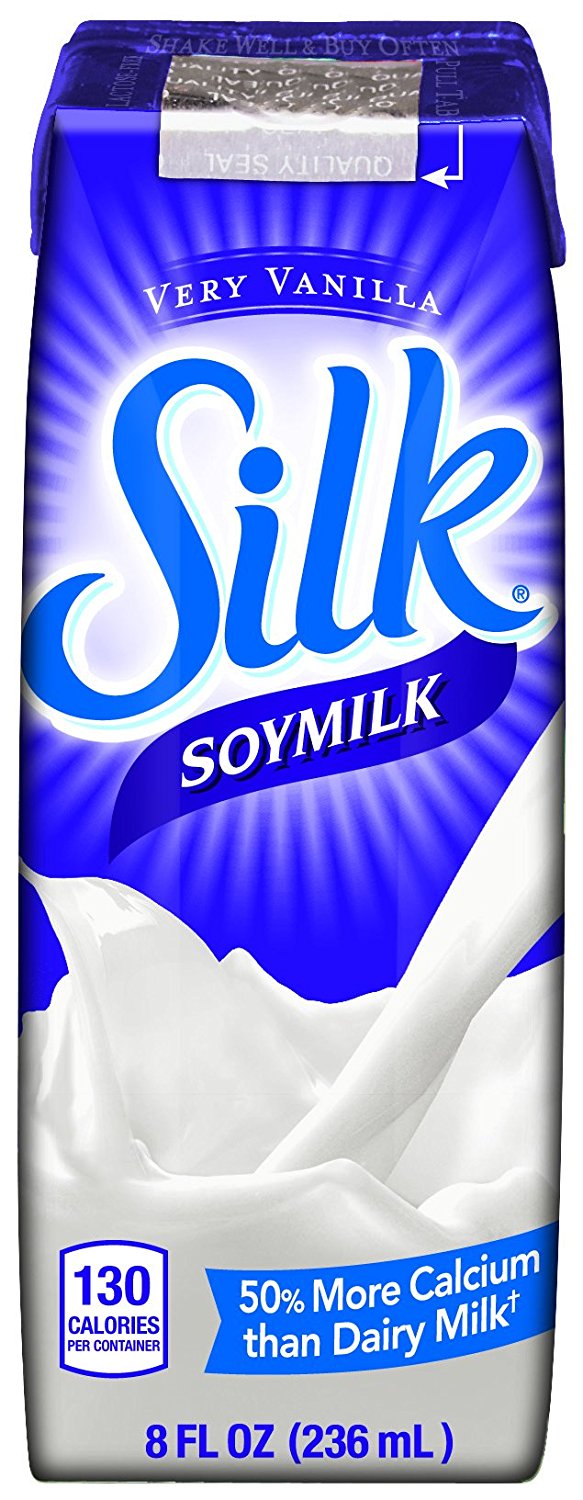}
    \vspace{2mm} \end{minipage}
        \hspace{-3mm}
    &
    \hspace{-3mm}
    \begin{minipage}{0.1\textwidth}
    \centering \vspace{2mm}
        \includegraphics[width=\linewidth, height=14mm, keepaspectratio]{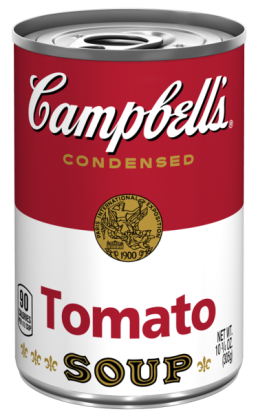}
    \vspace{2mm} \end{minipage}
        \hspace{-3mm}
    &
    \hspace{-3mm}
    \begin{minipage}{0.1\textwidth}
    \centering \vspace{2mm}
        \includegraphics[width=\linewidth, height=14mm, keepaspectratio]{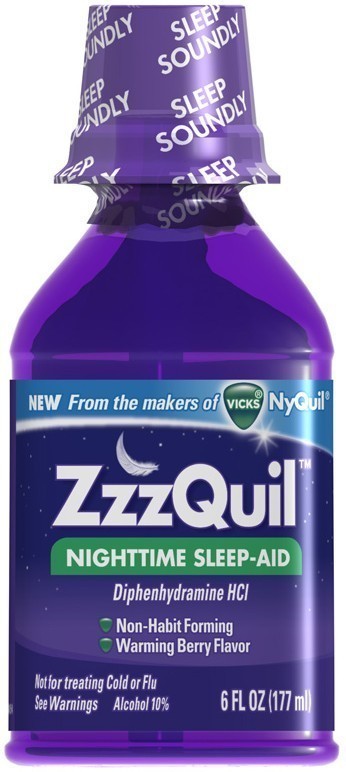}
    \vspace{2mm} \end{minipage}
    \\\hline
    \textbf{No regrasp} & 61 $\%$& 38 $\%$& 64 $\%$& 75 $\%$& 46 $\%$& 61 $\%$& 36 $\%$& 17 $\%$\\\hline
    \textbf{Tactile-based regrasp} & 85 $\%$& 61 $\%$& 78 $\%$& 93  $\%$& 63 $\%$& 83 $\%$&72 $\%$& 49 $\%$\\\hline
    \textbf{Relative Improvement} & 39 $\%$& 61 $\%$& 22 $\%$& 75 $\%$& 37 $\%$& 36 $\%$& 100 $\%$& 188 $\%$\\\hline
  \end{tabular}
  \label{table:regrasp_accuracy}
  \vspace{-3mm}
\end{table*}

\subsection{Comparison with heuristic baseline: centroid-centering} 
In an effort to contrast the performance of our proposed policy, we compare it to a tactile-based heuristic  approach: centroid-centering. This regrasp strategy computes the average centroid of the tactile imprints and adjusts the gripper position to relocate the computed centroid at the center of the tactile sensor. We applied this heuristic to both a known and an unknown object (\textit{glucose bottle}, \textit{tomato can}). For both objects the accuracy averaged over 100 grasps when centering the centroid (65$\%$, 53$\%$) improves the performance over open loop grasping (58$\%$, 36$\%$), but is substantially lower than using our grasp quality based policy (83$\%$, 72$\%$). We hypothesize that the superior performance of our policy over hand-crafted heuristics is due to the fact that the quality of the grasp is highly dependant on fine contact details that only a wholistic use of the tactile image can exploit.

exploits fine details in the tactile imprints to determine the quality of the grasps. 

\section{Discussion}



    
This paper presents an approach to improve grasp robustness using tactile feedback. We first learn a tactile-based grasp quality metric. The model is self-supervised, and parametrized using a deep convolutional neural network trained with 2800 tactile images from 12 different objects automatically labelled with their corresponding grasp qualities. The grasp quality metric is leveraged to design a reactive policy that makes local adjustments using tactile sensing. The regrasp policy acts by simulating tactile imprints in the vicinity of the initial grasp and selecting the ones with the highest predicted quality score. Results on a test set of 8 objects show that enabling tactile feedback yields average relative improvements of the accuracy of 70$\%$ over open-loop grasping. 

\begin{figure}[h]
\centering
{
		\includegraphics[width=\linewidth]{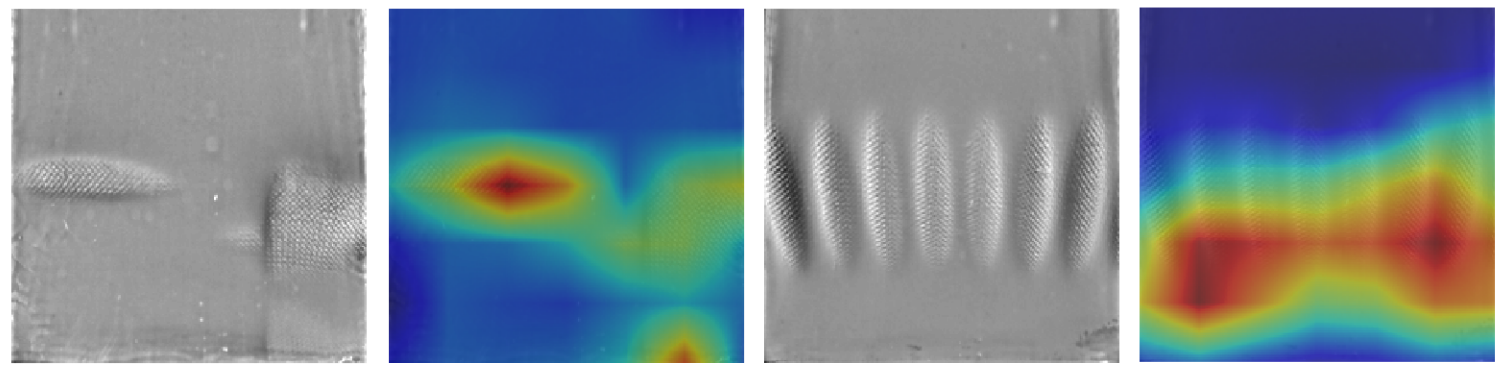} %
}
\centering
\caption{\textbf{Tactile images and CAM maps.} We observe that the regions in the tactile imprints that CAM finds more relevant when predicting grasp quality are highly correlated with the high intensity zones of the contact areas. } \label{fig:CAM}
\end{figure}

\myparagraph{What is the network looking at?} To understand what part of the tactile image is used by the network to predict grasp quality, we make use of Class Activation Mapping (CAM) \cite{zhou2015}, a computer vision technique that localizes the regions of an image that are relevant to the output of a neural net. Figure~\ref{fig:CAM} shows two examples of a tactile imprint and the corresponding CAM region. Following intuition, the areas of the tactile image that are more discriminative are highly correlated with the contact areas observed in those images. CAM pays special attention to those regions with high contrast suggesting that the grasp quality is strongly related to the degree of pressure.


\myparagraph{Limitations and future work.} One clear limitation of the presented tactile-based approach is the local nature of the considered regrasps. An interesting line of future work is to address this issue by considering the 3D geometry of objects and simulating more global tactile transformations. Another line of improvement is related to the discontinuous nature of our policy that has to leave the object before performing a regrasp. Allowing for continuous readjustments would better resemble the way humans manipulate objects and improve the smoothness of the regrasps. Finally, to further improve grasping performance, we are interested in extending the idea of planning grasp adjustments by efficiently combining tactile and visual information. 

\bibliographystyle{IEEEtranN}
{\footnotesize\bibliography{references}}

\end{document}